\documentclass[10pt,journal,compsoc]{IEEEtran}
\usepackage{epsfig}
\usepackage{amsmath,amssymb,amsfonts}
\usepackage{algorithmic}
\usepackage{graphicx}
\usepackage{textcomp}
\usepackage{xcolor}

\begin{document}

\title{A large-scale operational study\\ of fingerprint quality and demographics}

\author{Javier~Galbally\textsuperscript{1},
        Aleksandrs~Cepilovs\textsuperscript{1},
        Ramon~Blanco-Gonzalo\textsuperscript{2},
        Gillian~Ormiston\textsuperscript{3},
        Oscar~Miguel-Hurtado\textsuperscript{2},
        and~Istvan~Sz.~Racz\textsuperscript{2}
\IEEEcompsocitemizethanks{\IEEEcompsocthanksitem \textsuperscript{1}Research and Development Team, \textsuperscript{2}Biometrics Team, and \textsuperscript{3}Shared Biometric Matching Service Team,
European Union Agency eu-LISA, Tallinn (Estonia)/Strasburg (France).\protect\\
E-mail: javier.galbally@eulisa.europa.eu}
\thanks{Manuscript received October 2024}}

\markboth{Arxiv cs.CV 2024}%
{Shell \MakeLowercase{\textit{et al.}}: Bare Demo of IEEEtran.cls for Computer Society Journals}



\IEEEtitleabstractindextext{%
\begin{abstract}
Even though a few initial works have shown on small sets of data some level of bias in the performance of fingerprint recognition technology with respect to certain demographic groups, there is still not sufficient evidence to understand the impact that certain factors such as gender, age or finger-type may have on fingerprint quality and, in turn, also on fingerprint matching accuracy. The present work addresses this still under researched topic, on a large-scale database of operational data containing 10-print impressions of almost 16,000 subjects. The results reached provide further insight into the dependency of fingerprint quality and demographics, and show that there in fact exists a certain degree of performance variability in fingerprint-based recognition systems for different segments of the population. Based on the experimental evaluation, the work points out new observations based on data-driven evidence, provides plausible hypotheses to explain such observations, and concludes with potential follow-up actions that can help to reduce the observed fingerprint quality differences. This way, the current paper can be considered as a contribution to further increase the algorithmic fairness and equality of biometric technology.
\end{abstract}

}

\maketitle


%

\vspace{0.1cm}

\begin{quote}
\centering{``\emph{It's not the size of the dog in the fight, it's the size of the fight in the dog}'' - Mark Twain}
\end{quote}

\vspace{0.1cm}


\section{Introduction}
\label{sec:intro}

\IEEEPARstart{F}{ollowing} the evidence collected in different studies regarding the accuracy variability of fingerprint recognition technology with respect to age \cite{Galbally19ageingFps,haraksim19PRfpGrowth}, gender \cite{sharma21fpsSexEstimation} and finger-type \cite{Gafurov10fingerTypePerformance,galbally22IWBF}, the biometric community seems to agree that there exists a certain degree of bias in current fingerprint-based systems for different demographic groups. However, with the exception of a few studies, this inconsistency in the recognition rates has been mainly observed on small-to-medium databases under laboratory conditions and, therefore, it is difficult to quantify to what extent this bias may translate to large-scale systems working under real operational conditions.

The main objective of the present study is to contribute to close the existing gap on the availability of evidence on large-scale operational data where fingerprint bias across age, gender and finger-type can be observed. The data employed in the project was acquired at consulates all around the world for the issuing of visas, using the current most extended technology for scanning fingerprints (i.e., 500dpi touch-based optical scanners). Based on these data, the fundamental question addressed in the study is: assuming that the amount of identity-related information present in natural fingerprints is not linked to a specific demographic group, why do some segments of the population consistently present higher error rates when utilising fingerprint recognition technology? Or, put in other words, is it a valid assumption that natural fingerprints of different demographic groups intrinsically contain, from birth, an equivalent amount of identity-related information that can be leveraged by automated systems to recognise individuals?

In order to provide a plausible answer to this difficult query, other related topics also considered in the article are: Do fingerprints coming from men contain more information than those coming from women? Adults’ fingerprints comprise more information than those of young children or elders? Why do each of the fingers (including the thumb) of the hand provide different accuracy performance in fingerprint recognition systems?

The work tackles all these issues from a fingerprint quality perspective following a three-step approach: 1) determine if, indeed, there is a bias in fingerprint quality according to: age, gender and finger-type; 2) provide hypotheses that can reasonably explain the observed bias; 3) suggest a viable course of action to correct and/or minimise the observed bias.

The analysis and results presented in the paper can bring yet further insight into the key area of fingerprint quality from a new perspective, not considered in research publications to date, bridging one of the few gaps still existing in the field. As such, the conclusions drawn from the work, can bring huge value to different actors involved in the design, development and deployment of fingerprint-based operational systems:

\begin{itemize}
\item Industry producing fingerprint readers: provide them insight for the design of improved ergonomics, enhanced usability of fingerprint acquisition scanners and help them improve the technology behind fingerprint sensing.

\item Algorithm/application developers: give them some guidance for the development of improved recognition strategies which may take into account demographic features to optimise the performance for different segments of the population.

\item End-users in the domains of border management and law enforcement: help them develop guidelines and best practices for different applications/scenarios regarding the decision-making process with respect to different demographic groups.
\end{itemize}

The rest of the paper is structured as follows. After a brief discussion on fingerprint quality provided in Sect.~\ref{sec:fpQ}, the experimental framework, including the database and the quality metric which are the basis for the evaluation, are described in Sect.~\ref{sec:framework}. Results, together with the most relevant works in the literature related to each of the factors considered in the paper, are discussed in Sect.~\ref{sec:fpsgender} (gender), Sect.~\ref{sec:fpsage} (age) and Sect.~\ref{sec:fpstype} (finger-type and handedness). Finally, conclusions are drawn in Sect.~\ref{sec:conc}.

\section{A brief reflection on fingerprint quality}
\label{sec:fpQ}

It is nowadays a well-established and accepted fact among the biometric community, that quality of biometric samples is, without a doubt, the primary factor impacting the accuracy and overall performance of biometric recognition systems. Therefore, any variability observed in the accuracy of fingerprint-based applications with respect to demographic features, can be mainly linked to discrepancies in the quality level of fingerprints produced for each of the demographic groups considered.

From the perspective of automated recognition systems, ultimately, fingerprint quality can be directly connected to the amount of usable identity-related information that can be reliably and consistently extracted from the digital representation of a fingerprint (which in the vast majority of cases is a 2D image). In turn, this amount of information, will determine to what extent the fingerprint is unique and, therefore, will provide high accuracy in recognition tasks. As such, coming full circle, quality becomes an estimation or a prediction of accuracy, as defined by the ISO/IEC 29794-1:2016 standard.

There is a key concept that should be highlighted and that can easily remain somewhat hidden in the previous paragraph, within the adjective ``\textit{usable}''. It is important to distinguish between 1) the amount of identity information contained in the natural fingerprint, and, from that naturally built-in intel, 2) the amount that is captured by the digital representation of the fingerprint and that can be extracted (i.e., is ``\textit{usable}'') by automated systems. Fundamentally, it is this digitally \textit{extractable} and \textit{usable} information that defines the quality of a fingerprint and its potential to be used in recognition tasks.

As such, it is of great consequence to minimise the amount of information that is lost in the conversion process from the natural fingerprint to its digital representation. This translation ``from the natural world to the digital domain'' is accomplished at the time of acquisition by the fingerprint reader. It follows that fingerprint quality, and therefore also the eventual accuracy provided by automated recognition systems, is mainly determined at the time of acquisition. Once a fingerprint image is captured, it is difficult to enhance its quality or to improve the amount of usable information in the digital representation of the fingerprint. Known factors that determine this quality level at the time of acquisition can be mainly categorised in four groups: environmental related (e.g., temperature, humidity or lighting); finger related (e.g., dry skin, dirty or oily skin); reader related (e.g., resolution, ergonomics, sensing technology, post-processing required); behavioural, related to the human-reader interaction (e.g., positioning of the fingers, pressure applied in case of touch-based readers).

While, it is difficult to have any control over finger related conditions, in the case of supervised controlled scenarios, there do exist a broad range of actions that can be triggered in order to exert a positive effect on acquisition factors concerning the environment, the reader and the behaviour of the individual.

\section{Experimental framework}
\label{sec:framework}

\subsection{Database}
\label{sec:framework:DB}

The database used for the present study was captured in the span of three months, between March and May 2022, in the context of a pilot project carried out jointly between the Swedish Government through their Migration Agency and eu-LISA. The collaborative effort was focused on the improvement of the processes involved in the issuing and control of visas for non-EU citizens entering the Schengen area and, in particular, on the testing of tools for biometric quality assessment.

The database is formed by a total 15,942 different 10-print digital records produced by as many individuals, that is, it contains a total 159,420 fingerprint samples. Individuals come from 34 different non-EU countries around the world. Fingerprints were captured in 115 different locations with specific designated stations for visa issuing purposes (typically consulates). Therefore, fingerprints were captured in office-like scenarios, and the process was conducted by operators with experience and instruction in the field of fingerprint acquisition.

All fingerprints were captured using the same FBI-certified touch-based 500 dpi optical scanner (Cross Match Patrol ID). All fingerprint images are flat (i.e., not rolled). For each 10-print record three different images were captured, following the typical sequence 4-4-2, that is: slap of the right hand (all four fingers acquired simultaneously), slap of the left hand (all four fingers acquired simultaneously) and lastly the two thumbs acquired at the same time.

In case of low quality, fingerprints were reacquired up to three times, and the best individual quality score for each finger was kept in the final composite record.

For each fingerprint, the meta-data available in the database is: gender; age; country of origin; code of the station where it was acquired; finger-type (left or right; little, ring, middle, index or thumb); cycle number of the acquisition attempt (i.e., 1, 2 or 3).

\begin{figure}[t]
\centerline{\epsfig{figure=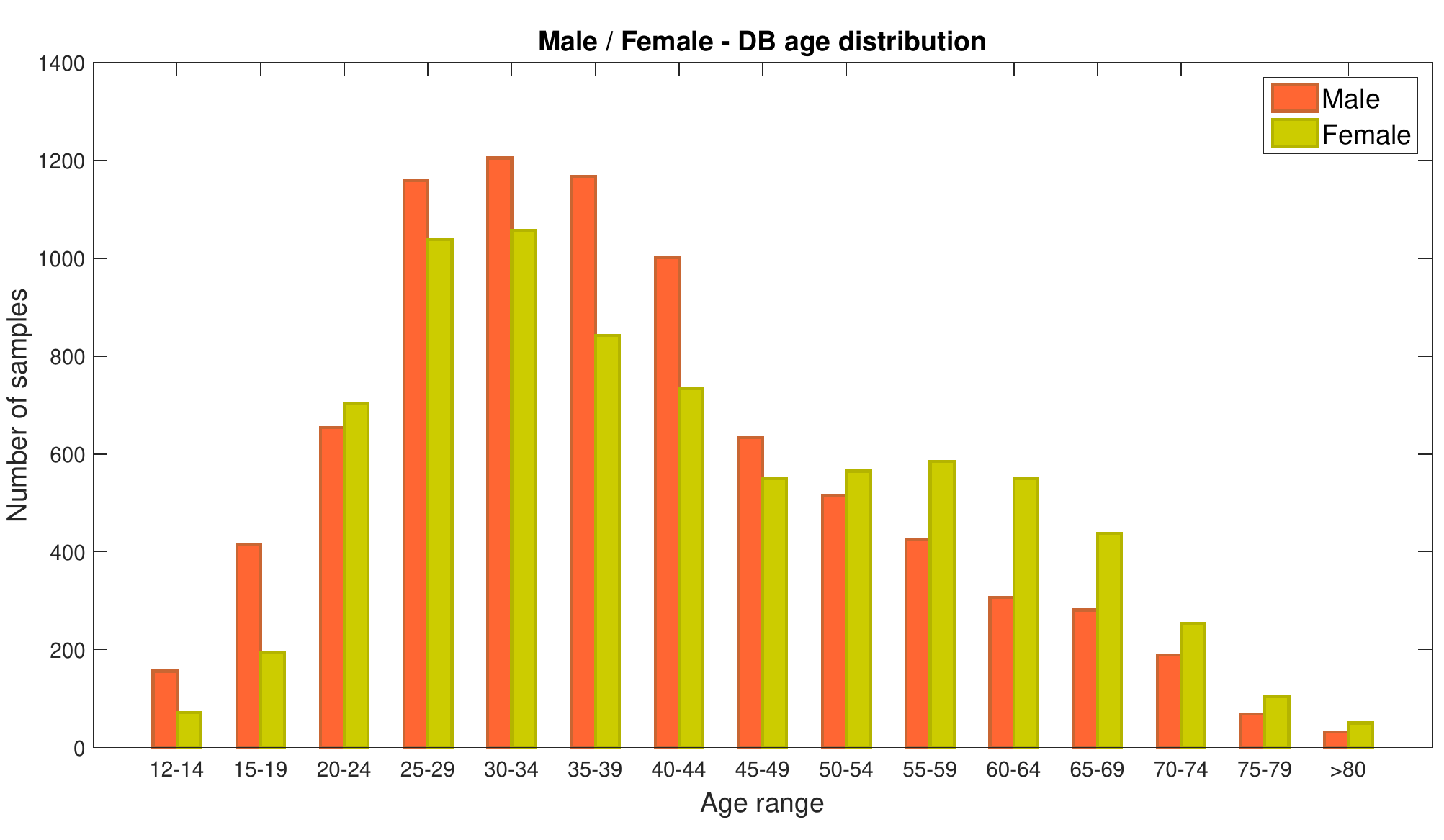,width=0.99\linewidth}}
\caption{Age distribution in the experimental database according to sex.}
\label{fig:ageDBdist}
\end{figure}

The gender ratio men/women in the database is 8206/7736 (i.e., 52/48\%). The bar graph in Fig.~\ref{fig:ageDBdist} shows the age distribution in the database, using colour code (light red for men and light gold for women), to distinguish between the number of women and men for each age group. All subjects are above 12 years of age, as that was the minimum age for compulsory fingerprinting for the issuing of visas at the time the data was acquired.

Due to data protection reasons, for the experimental evaluation we did not have direct access to the fingerprint images, which were retained by the Swedish authorities, owners of the operational data. The analysis was carried out solely based on: 1) the quality scores extracted by the Swedish colleagues from the fingerprint samples; 2) the metadata corresponding to each sample in the database.

While the list of countries and the number of subjects per country is available, this information is not disclosed in the present paper, as we believe it has a negligible impact in the final outcome of the experiments.

\subsection{Quality measure: NFIQ-2}
\label{sec:framework:NFIQ2}

For the experimental evaluation, NFIQ-2 was used as the tool to assess quality. NFIQ-2 is a system- and vendor-agnostic fingerprint quality measure which is enforced as reference implementation of the ISO/IEC 29794-4 standard on fingerprint quality. The source code, which is publicly available, was initially developed as an initiative of US NIST in response to the need of having reliable quality assessment tools dissociated from specific vendors \cite{NFIQ2code}. Currently, the project is updated and maintained by the ISO SC 37 Working Group 3. The NFIQ-2 quality measure has been independently evaluated in numerous occasions, showing very high performance across recognition systems, and has nowadays been adopted by the biometric community as the \textit{de facto} standard to set base results to which other quality measures are compared. The current version of NFIQ-2 is trained on flat fingerprints of 500 dpi resolution, captured with optical devices, that is, the same category of fingerprint images contained in the experimental database used in the present work. There is an ongoing project to extend NFIQ-2 to assess also the quality of rolled fingerprints and of other resolution values.


\section{Results}
\label{sec:results}

As a general caveat, when analysing the results it is important to bear in mind that, as was described in Sect.~\ref{sec:framework:DB}, fingerprints in the experimental DB were obtained in a supervised highly controlled scenario, with a re-capture policy to be followed by operators in case the acquired images did not reach the desired quality threshold.

This means that the impact of behavioural or environmental factors (e.g., positioning of fingers, pressure applied, cleanliness of fingers) in the resulting quality of fingerprint images is largely minimised, unlike other uncontrolled and unsupervised scenarios (e.g., the current trend of touchless applications for the self-acquisition of fingerprints with smartphone cameras). Consequently, given the specificities of the experimental DB, potential differences in the fingerprint quality levels across population groups should be mainly originated by flaws or lack of consistency in the overall system functioning, including both the reader and the subsequent post-processing steps of fingerprint images.

\subsection{Fingerprint quality and gender}
\label{sec:fpsgender}

\subsubsection{Works related to fingerprints and gender}
\label{sec:fpsgender:related}

Several works published in the scientific literature have studied the differences between female and male fingerprints, predominantly in the context of the development of sex detection algorithms \cite{sharma21fpsSexEstimation}. In a nutshell, the key finding of all this previous research can be summarised as: there exists a measurable difference between female and male fingerprints in terms of overall size and ridge width, which results, ultimately, in a difference in the ridge density. In essence, it has now been soundly established that female fingerprints present, on average, thinner ridges contained in a smaller surface, yielding a higher ridge density that can be exploited to classify fingerprints according to sex, a useful feature, for instance, in criminal investigations \cite{spanier24fpsGenderForensics}.

\subsubsection{Results: fingerprint quality and gender}
\label{sec:fpsgender:results}

In the present study we analyse the potential divergence of men and women fingerprints from a different perspective: quality. The objective is to determine whether or not fingerprints present different quality levels according to sex attributes, which would, in turn, trigger possible biases in the performance of fingerprint-based recognition systems for men and women.

\begin{figure}[t]
\centerline{\epsfig{figure=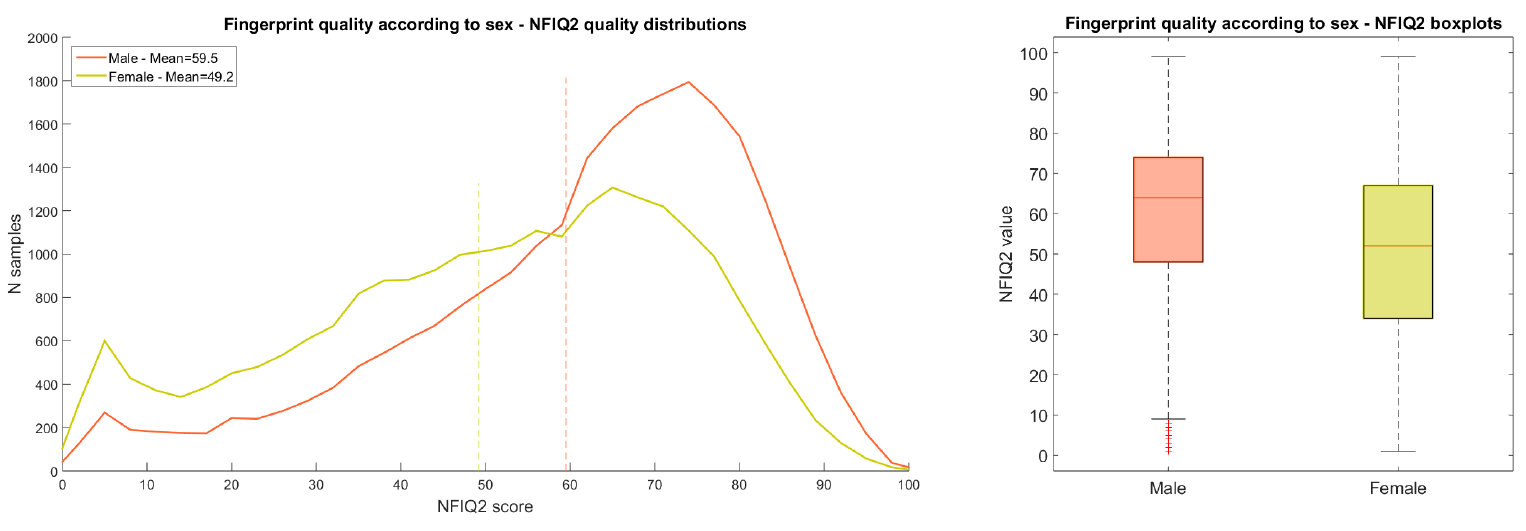,width=0.99\linewidth}}
\caption{NFIQ-2 quality distributions for men and women (left) and the corresponding box plots for those distributions (right). On each box, the central mark indicates the median, and the bottom and top edges of the box indicate the 25th and 75th percentiles, respectively. The whiskers extend to the most extreme data points not considered outliers, and the outliers are plotted individually using the '+’ marker symbol.}
\label{fig:sexResults}
\end{figure}

Fig.~\ref{fig:sexResults} shows, on the left, the NFIQ-2 quality distribution for all the fingerprints in the experimental database for men (light red) and women (light gold). The box-plots corresponding to the two distributions are depicted on the right. In each box, the central mark indicates the median, and the bottom and top edges of the box indicate the 25th and 75th percentiles, respectively. The whiskers extend to the most extreme data points not considered outliers, and the outliers are plotted individually using the '+’ marker symbol.

\textbf{OBSERVATION}. Results show a significant difference in the quality levels of female and male fingerprints, not only on the average value 59 to 49, but also on the consistency of the quality produced, with a noticeably lower variance for men fingerprints (i.e., it is rare that men produce fingerprints of lower quality).

\textbf{DISCUSSION}. As was presented in Sect.~\ref{sec:fpQ}, there is a direct link between fingerprint quality and usable amount of information within the fingerprint. A surprisingly extended misconception in the early days of the development of automated fingerprint recognition technology, among the general population, was that smaller fingerprints would contain less information than larger fingerprints and, therefore, they would be worse suited for recognition purposes, resulting in lower quality levels and higher error rates. Such erroneous belief, would be in some cases referenced to explain the lower accuracy observed for fingerprints coming from women and also young children, compared to those of adult males.

However, if such ``size vs information'' correlation assumption was to be proven correct, it would put at risk the credibility and reliability of fingerprint recognition technology as a whole, as it would go against one of its basic founding principles: the permanence of fingerprints. Fingerprints are formed during the fetal stage of pregnancy, being set by week 17 \cite{kucken05fpsFormationFetus}. After their formation, fingerprints have been shown empirically through uncountable experimental studies, to remain invariant in terms of the ridge structure, throughout the lifetime of individuals (with the exception of damage caused by external environmental conditions, such as scars). Different works have provided solid data evidence that supports that the only inherent internal change to fingerprints is the homomorphic growth during childhood \cite{Galbally19ageingFps}. However, both the ridge pattern and minutiae points remain stable. This means that, when we are born, our fingerprints are indeed smaller, but already contain all the identity related information that will allow automated systems to accurately perform recognition of individuals during adulthood.

Based on the rationale above, it follows that the key factor defining the amount of information contained in fingerprints, is not the absolute size of the fingerprint itself, but the density of the ridge pattern. Smaller fingerprints also present thinner ridges and valleys and, as a result, can comprise the same amount of information as larger fingerprints in a smaller surface. That is, the information density is higher. It goes without saying that, inherently, in their natural state, some fingerprints will contain more information than others, as was already statistically shown in the famous ``Doddington’s zoo'' work over two and a half decades ago \cite{doddington98zoo,yager10menagerie}. However, this amount of identity-related information, or level of uniqueness, should bear no interdependence with the size of the fingerprint itself.

Indeed, the forensic and biometric community have now long shown that fingerprint size does not necessarily correlate to the amount of information contained within them. While there may not be specific studies directly addressing whether smaller fingerprints contain less information, the consensus in the scientific community is that the uniqueness of fingerprints is primarily determined by the detailed ridge patterns and minutiae, rather than their overall size.

The discussion presented so far leaves one open question: provided that the size of fingerprints is not the inherent primary factor in determining fingerprint quality and, in turn, also fingerprint accuracy: why do smaller fingerprints coming from women consistently present worse quality levels, and therefore also perform worse, than those of adult males?

On the one hand, current standard Automated Fingerprint Identification Systems (AFIS) work at a resolution of 500dpi, which may not be enough to properly capture the structure of smaller fingerprints with higher ridge density such as that of women and children. If the resolution is indeed insufficient, after acquisition, it is not possible to recover the information potentially lost in the translation stage performed by the scanner, from the fingerprint in its natural stage to its digital representation (i.e., fingerprint image sample).

On the other hand, even in the cases where 500dpi is enough to properly capture all the information contained in smaller fingerprints, the image post-processing steps that are typically applied to the sample in order to generate the final template may be optimised for larger fingerprints, performing worse with fingerprints of finer details (e.g., segmentation of the region of interest, normalisation, filtering and computation of the orientation field, binarisation, ridge thinning, minutiae detection and extraction).

\textbf{HYPOTHESIS}. The quality difference between women and men fingerprints, is not due to intrinsic properties of the fingerprints themselves, but to current fingerprint recognition technology being designed to acquire and process, in a more accurate fashion, fingerprints of the size and ridge density of adult men.

\textbf{SOLUTIONS}. In order to test if the previous hypothesis holds, two course of action may be taken. The combination of these two corrective measures should show an improvement in the performance of female fingerprints, bringing it closer to that achieved by male fingerprints.

First, the resolution of scanners can be increased to 1000 dpi, which is in fact the standard resolution employed by forensic examiners today in the manual comparison of fingerprints and finger-marks. The sensing technology to produce higher resolution readers is already available, and it would ensure the minimisation of the information lost during acquisition. Historically, the resolution of 500 dpi was selected as it offered a good compromise between the fidelity with which the natural fingerprint is acquired and the size of the resulting digital image \cite{NIJ11fpsSourcebook}. However, nowadays, with the continuous increase both in storage capacity and in computational speed of digital systems, the size of the captured images is no longer a limitation factor, and preference should be given to the acquisition of more detailed images of higher quality. We could argue that, currently, digital technology (sensing, storage and processing) has reached a point where there is no reason not to capture fingerprints of higher resolution. Eventually, it is always better to reduce the amount of information acquired, should it be redundant, than to estimate or interpolate missing information. In other words, following one of the primary principles of data science: the more data, the better.

Second, fingerprint recognition systems could use image post-processing algorithms specifically trained and tailored to a given fingerprint size and ridge density values. That is, systems may deploy ``ridge density-specific'' (i.e.,``sex-specific'' or ``age-specific'') algorithms for the processing of fingerprints. These adaptative methods would help equalising the performance of fingerprints with different information density values (such as those from women or children).

\subsection{Fingerprint quality and age}
\label{sec:fpsage}

\subsubsection{Works related to fingerprints and age}
\label{sec:fpsage:related}

Several works have addressed in the scientific literature the dependency of fingerprint quality with respect to age. Most of these works focus on the analysis of the performance of children fingerprints. Among these contributions centred on youngsters,

two works that stand out as the most complete are \cite{jain17tifsFPsChildren} and \cite{haraksim19PRfpGrowth}. The first of these studies \cite{jain17tifsFPsChildren}, was published in 2017 by a team from the Michigan State University. The evaluation was carried out on fingerprints from 309 children aged between 0 and 5 years, captured using both a standard 500ppi resolution reader, and a customed high-resolution 1270ppi reader. The objective was to address some open questions at the time such as: 1) Do fingerprints of young children possess the salient features required to uniquely recognize a child? 2) If so, at what age can a child’s fingerprints be captured with sufficient fidelity for recognition? 3) Can a child’s fingerprints be used to reliably recognize the child as he ages? The main conclusion of the work was that, in fact, children fingerprints contain all the identity-related information that allows for accurate and consistent recognition, provided that the acquired images are of sufficient quality and resolution.

These results were confirmed in the second study mentioned above carried out in 2019 by a team of scientists working at the European Commission’s Joint Research Centre (JRC). This work was performed on a database of almost 70K pairs of fingerprints coming from children aged 5-16, that were captured at different points in life with a difference of between 1 to 7 years \cite{haraksim19PRfpGrowth}. The experiments showed that the amount of identity information contained in the fingerprints remained constant during childhood growth, being independent of the fingerprint size. The study verified for the first time that the displacement of minutiae points during this period of life due to growth follows an isotropic model, that is, the displacement is invariant to the distance between the minutiae and the centre of the fingerprint and to the actual location of the minutiae with respect to this centre. Based on this finding, the authors developed a growth model capable of compensating such displacement in order to reduce the ageing effect on fingerprint recognition systems (i.e., the decrease in accuracy when the time separation between the two compared fingerprint impressions increases).

From a general perspective, not focused solely on fingerprints coming from children, the most comprehensive study on the evolution of fingerprint quality and accuracy through the full human life-span was carried out in 2017 by the same JRC research team that published the study described in the previous paragraph \cite{haraksim19PRfpGrowth}. The experiments were carried out over almost 500K fingerprints, coming from individuals aged 0-25 and 65-98 years, acquired under operational conditions with standard 500ppi optical touch-based scanners for the purpose of issuing ID national cards \cite{Galbally19ageingFps}. The most salient observations from the experiments were that: 1) from a quality perspective, children fingerprint impressions show better quality than those of the elderly; 2) fingerprint quality increases rapidly between 0 and 12 years of age, where it stabilises, remaining fairly constant during adulthood until 40-45 years of age, where it starts decreasing linearly.

It should be noted that, in the JRC study, due to lack of fingerprint data in the age range 26-64, the behaviour of fingerprint quality during that period was provided as an estimation, based on linear fitting of the available data, rather than an experimental observation. As stated in that work ``\textit{given the limited amount of data available for adults from an age-wise perspective, covering ages 18-25, this assumption regarding the stable behaviour of fingerprint quality for adults until approximately 45 years of age, where it starts decreasing linearly, should still be confirmed on a set of data covering the age range 26-64.}''

Even though undoubtedly of great value due to the amount of data (500K fingerprints) and the nature of these data (acquired in a real operational scenario), unfortunately the JRC study also lacked sex-related information, so that potential differences between the quality of men and women fingerprints could not be analysed.

\subsubsection{Results: fingerprint quality and age}
\label{sec:fpsage:results}

In the present work we address some of the questions left open in the JRC study \cite{Galbally19ageingFps}, in particular: 1) confirm (or reject) the assumption that fingerprint quality remains stable through adulthood until around 45 years of age, when it starts decreasing linearly; 2) analyse if there exists any significant difference between the evolution of women and men fingerprints quality through life.
With this two-fold objective in mind, Fig.~\ref{fig:age+sexResults} shows, using boxplots, the evolution of fingerprint quality, both for men (light red) and women (light gold) between 12 and over 80 years of age. Different observations can be extracted from the results presented in this figure:

\begin{figure*}[t]
\centerline{\epsfig{figure=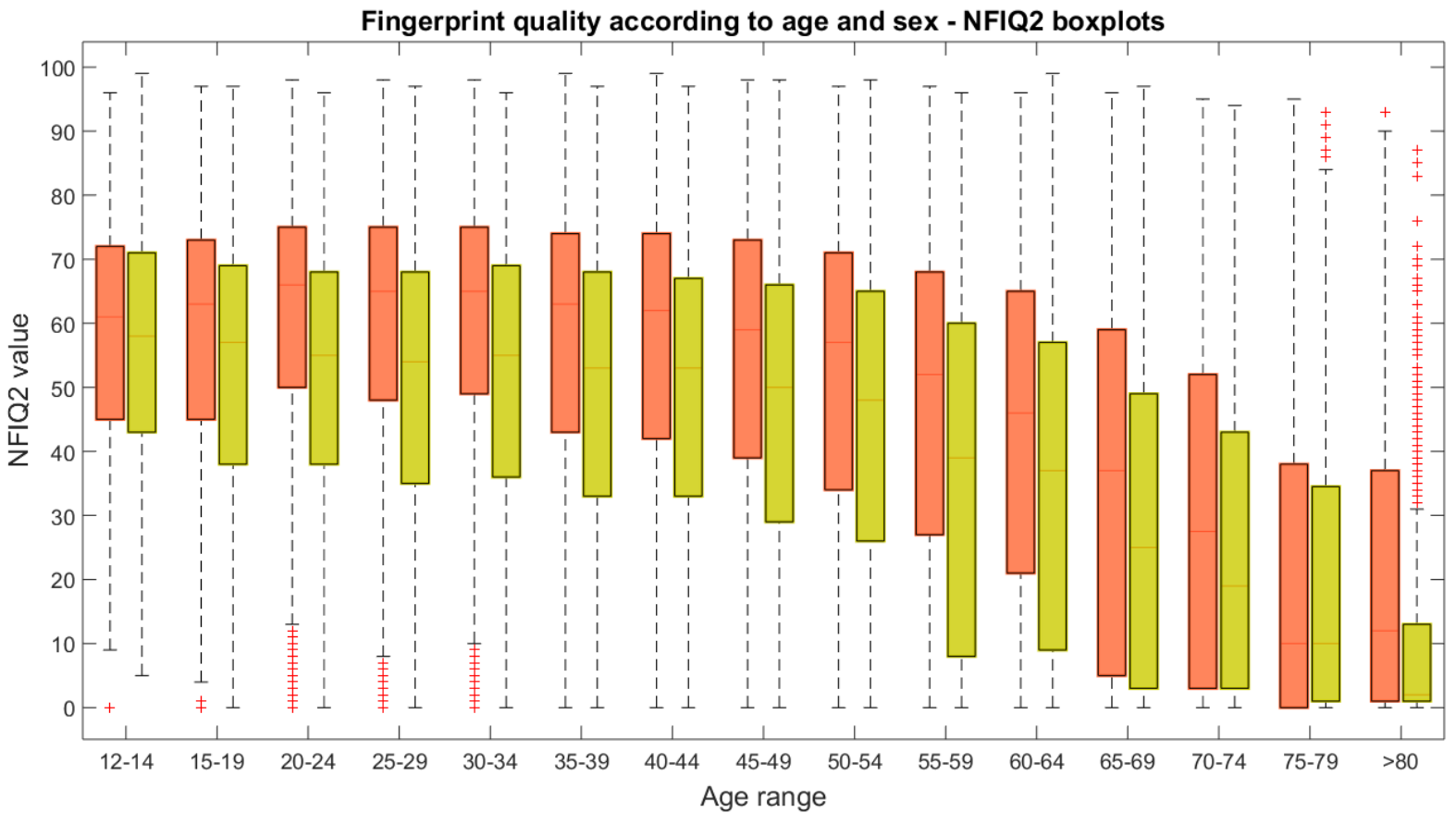,width=0.9\linewidth}}
\caption{Boxplots representing the fingerprint quality for the different age groups present in the experimental database, separated by sex (men, light red, and women, light gold). the central mark indicates the median, and the bottom and top edges of the box indicate the 25th and 75th percentiles, respectively. The whiskers extend to the most extreme data points not considered outliers, and the outliers are plotted individually using the '+’ marker symbol.}
\label{fig:age+sexResults}
\end{figure*}

\textbf{OBSERVATION 1}. As a first new finding, the results provide data-based evidence to confirm the hypothesis made in the JRC work \cite{Galbally19ageingFps}, and that still needed to be verified: fingerprint quality stays stable from 12 years of age, through the first part of adulthood, until around 45-50 years of age, when it starts decreasing linearly with age.

\textbf{OBSERVATION 2}. A second new piece of evidence with respect to previous literature, is that the quality of women fingerprints is consistently lower than that of men through the life period covered in the experiments (12 to 80 years of age). The quality difference remains fairly constant for all the age groups covered in the database, except for the case of the youngest subjects (12-14 years of age), where fingerprint quality appears to be almost analogous independently of gender, coinciding with the time period where the size difference between boys and girls is also the smallest. Whether or not this lower quality for female fingerprints is also observable between 0 and 12 years of age (period not covered by the data in the present study) still needs to be assessed.

\textbf{OBSERVATION 3}. The results also support what was already pointed out by the experiments in \cite{Galbally19ageingFps}, regarding the quality of fingerprints of 12-year-old children. The quality level at this young age is already equivalent to that of adults between 18 and 45, and higher than that of adults over 50.

\textbf{OBSERVATION 4}. The present study also provides further compelling evidence to confirm the observation already made in \cite{Galbally19ageingFps} regarding the potential issues that may arise in the processing of fingerprints coming from elders above 65 years of age, as a result of the very poor quality of their fingerprint impressions.

\textbf{DISCUSSION}. The findings of the study point out that the elderly can pose a significant challenge to fingerprint recognition systems, comparable, or even bigger, than children. This fact can have big practical implications. We should not forget that Europe has stated a commitment to ``\textit{the rights of the elderly to lead a life of dignity and independence and to participate in social and cultural life}'' \cite{EUParl00EUfundamentalRights}. This implies the need to take measures to ensure the inclusion of elders in every-day life and to guarantee their access to services available to the general population. The results presented in this section show that, given the deterioration of fingerprint quality at advanced points in life, there is a potential risk of age-based discrimination against elders due to increased rates of failure-to-capture or failure-to-enrol. We believe that this should be an important issue to be considered in the design of fingerprint recognition systems in order to avoid possible inter-generational inequality \cite{rebera12ethicalAgeing}.

Provided that for the elderly, as for adults, fingerprints size and ridge density remains basically invariable with age, the question that follows is: Why then, is fingerprint quality of elders so much lower than that of adults, both for men and women? Why do experiments show a constant linear decrease of fingerprint quality after approximately 50 years of age? The main difference in this case is not size and ridge density (as was the case for men and women), but skin condition. It has been shown in the specialised literature that, with age, and especially starting at around 45 years of age, skin gradually loses elasticity, firmness and also becomes drier, mostly due to the decrease of collagen \cite{carmeli03ageingHand}. This progressive deterioration of skin properties, together with other possible medical sufferings typical of older ages such as arthritis, hinder the acquisition of fingerprint impressions based on current touch-based optical scanners, as the interaction between the reader platen and the finger does not happen in the optimal expected manner.

\textbf{HYPOTHESIS}. The low quality of fingerprints coming from elders is related to the deficient interaction between the finger and touch-based optical scanners due to the gradual and constant degradation of skin condition through life, being especially noticeable after 45 years of age.

\textbf{SOLUTIONS}. Over the last decade there has been a big investment within the biometric community to develop a new generation of fingerprint touchless acquisition methods based both on specifically designed readers and on standard equipment such as smartphones \cite{priesnitz21touchlessFPsReview}. While much progress has been made in the field, touchless fingerprint recognition technology is still lacking behind in some aspects with respect to the traditional and better-established touch-based procedures, as it adds variability and less controlled conditions to the acquisition process \cite{orandi20NISTTRtouchlessFPs,libert19NISTTRtouchlessInteroperability}. However, for some use-cases such as the acquisition of elder fingerprints, that prove to be highly challenging for touch-based readers (as shown in the experiments reported in \cite{Galbally19ageingFps} and in the present work), the use of touchless scanners can provide an improvement with respect to the current state of play. Avoiding contact between the skin and the sensor would solve the quality issues derived from this poor interaction and should, in theory, improve the overall performance of fingerprint recognition systems for this segment of society.

Further experimental evaluation needs to be performed to verify this possible solution, on a dedicated database acquired both with touch-based and touchless technologies and including both adults (as control group) and elders.

\subsection{Fingerprint quality and finger-type}
\label{sec:fpstype}

\subsubsection{Works related to fingerprints and finger-type}
\label{sec:fpstype:related}

Only a few early studies, carried out on small sets of data, consider the analysis of quality and recognition accuracy from each finger individually. Furthermore, except for two of these research publications \cite{Gafurov10fingerTypePerformance,Michels10fingerCombinationsFpRec}, in general, existing literature only addresses the topic as a by-result of experimental evaluations with a different main focus. However, even if carried out on small sets of data captured \textit{ad-hoc} in laboratory conditions, all these pioneer studies already point out to the possibility that the quality level and performance of the images produced by each individual finger may vary quite significantly.

The two most relevant works from the state of the art, related to the current piece of research, were both published in 2010. In the first of these studies \cite{Gafurov10fingerTypePerformance}, researchers from the Gjovik University analysed the influence of finger types on fingerprint recognition performance, over a database containing all 10 fingers of 100 subjects. Fingerprints were captured individually (not slaps impressions) using six different scanners, five touch-based and one touchless. Their analysis confirmed for the first time following a rigorous scientific protocol, the general claim that was commonly made to that date, without a solid experimental basis, regarding the lesser accuracy of the little finger for recognition tasks.

In the second of the 2010 studies, the authors examine how fingerprint recognition systems can balance the speed of a single-print system with the robustness of a ten-print system by using a combination of fingers \cite{Michels10fingerCombinationsFpRec}. The goal of this research was to find the combination of fingers that provides the best trade-off between acquisition/verification speed and fewest comparison errors. For this objective, a database containing images of all 10 fingers from 70 subjects was used. It was found that the thumb, index, and middle fingers of both hands presented the highest quality scores and were, accordingly, also the fingers providing the best accuracy in recognition tasks.

Therefore, the two 2010 studies summarised above, coincide in their conclusions on the quality and accuracy of finger-types. They both showed that the little fingers present the worst performance of all fingers, while the best is reached using the thumbs and indexes. These observations were, to a large extent, further reinforced in a technical report by the US NIST in 2018 \cite{Fiumara18NISTDB300}.

\subsubsection{Results: fingerprint quality, handedness and finger-type}
\label{sec:fpstype:results}

Building upon the findings of these preliminary publications, in the current section we present the results of the first large-scale operational study of fingerprint quality based on individual fingers. The objective of the analysis is to determine if, based on the current most extended acquisition devices (i.e., 500dpi touch-based optical scanners), there is a difference in the quality level of each finger. That is, we want to give an answer to the question: are all fingers born equal (in terms of quality)? Or put in another way, traditionally, all fingerprints are treated the same in terms of fingerprint recognition, but should they? Do all fingers produce images that present the same discrimination potential? Are fingerprint samples produced by all fingers equally suited for personal authentication? Do all fingerprint samples possess the same amount of usable discriminative information independently of the finger that produced them?

\begin{figure}[t]
\centerline{\epsfig{figure=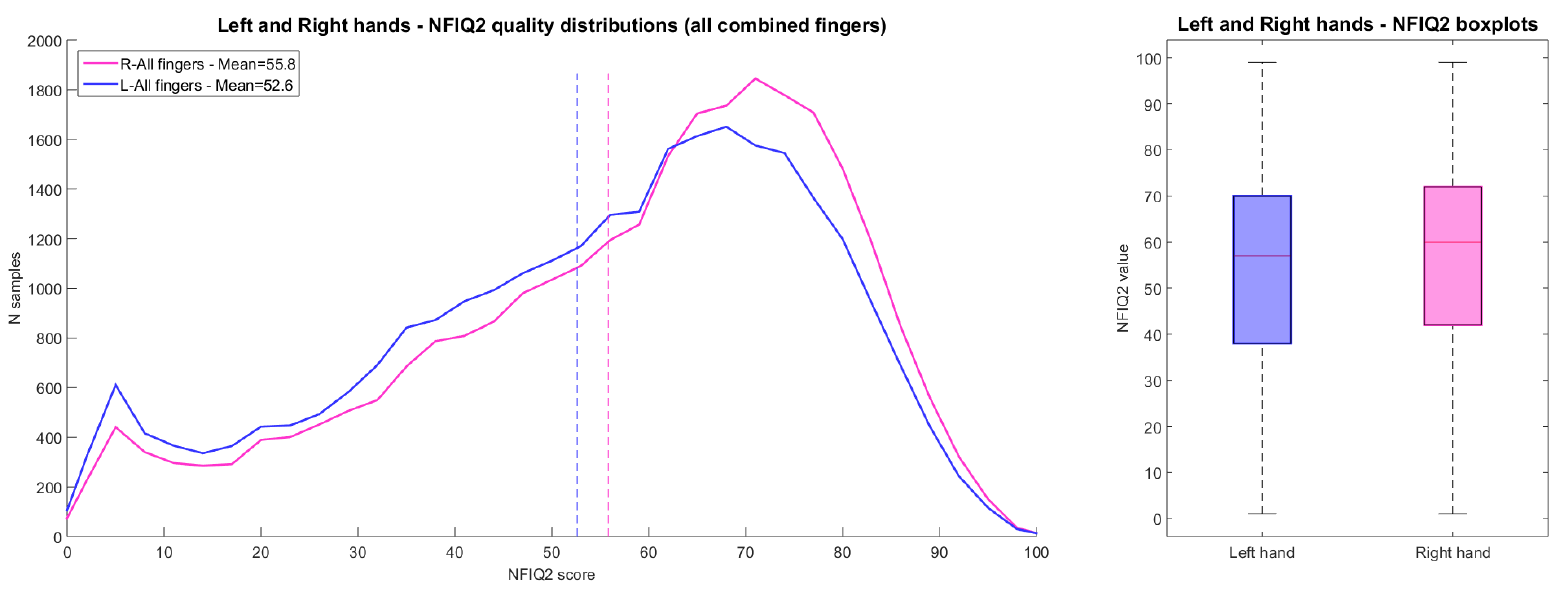,width=0.99\linewidth}}
\caption{NFIQ-2 quality distributions (left) for all fingers in the database separated by hand and the corresponding box plots (right) for those distributions.}
\label{fig:fingerTypeResults-hands}
\end{figure}

Fig.~\ref{fig:fingerTypeResults-hands} (left) shows the NFIQ-2 quality distribution for all the fingers combined of the left hand (blue) and of the right hand (pink). On the right, the box-plots corresponding to the two previous distributions are depicted.

\textbf{OBSERVATION 1}. From Fig.~\ref{fig:fingerTypeResults-hands} it can be observed that the right hand consistently provides better quality fingerprint images than the left hand.

\textbf{DISCUSSION}. We believe that this first observation is related to the handedness of humans. It is estimated that around 90\% of the world population is right-handed. As such, it is expected that individuals are more skilled to interact with the acquisition scanner using the right hand (their dominant hand) and, therefore, to provide better quality fingerprints.

\begin{figure*}[t]
\centerline{\epsfig{figure=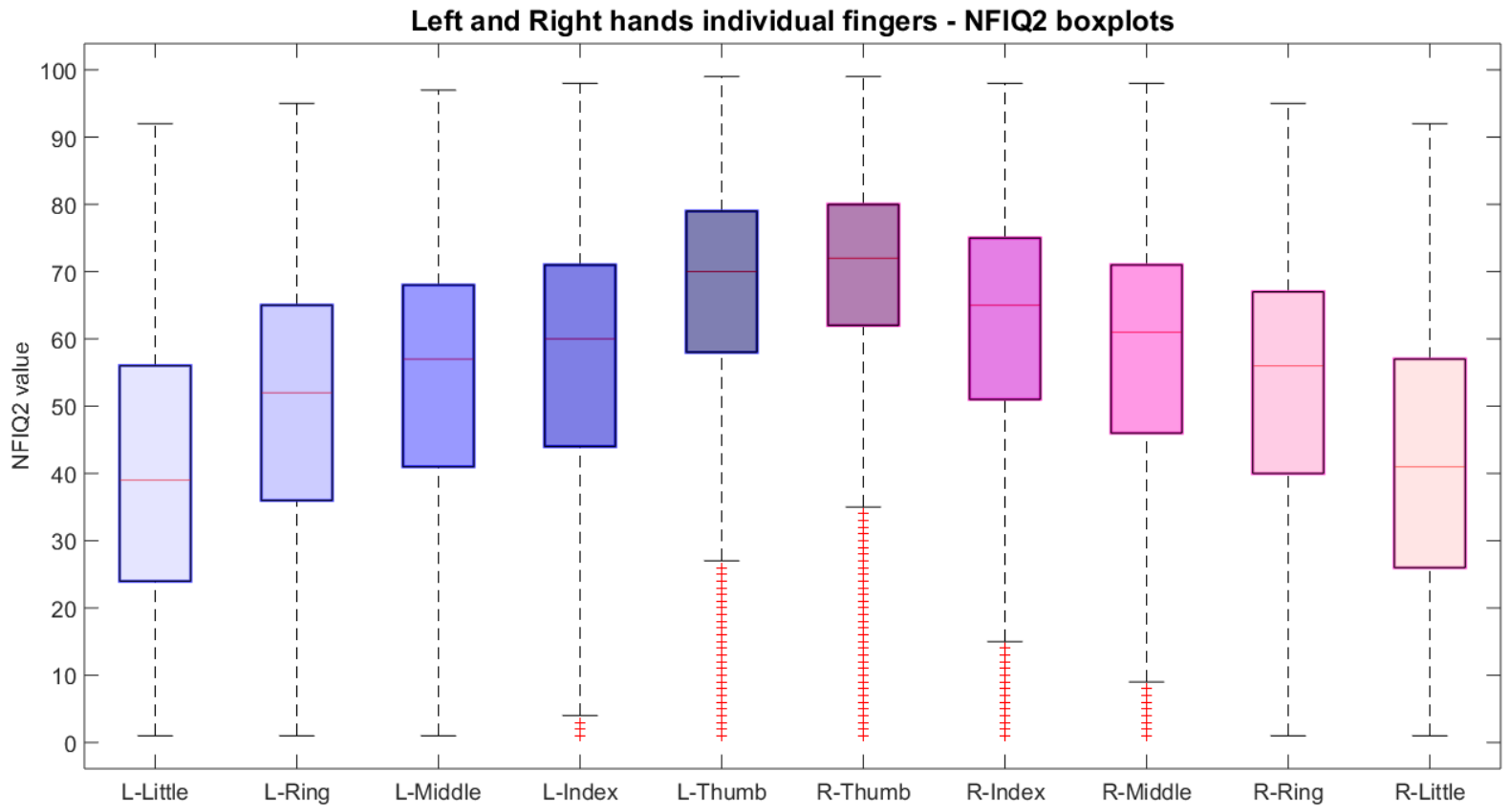,width=0.9\linewidth}}
\caption{Box-plots corresponding to the NFIQ-2 quality distributions per finger-type. For clarity, the box-plots follow the natural order of the fingers of both hands. The box-plots corresponding to fingers of the left hand are depicted in shades of blue, while those corresponding to fingers of the right hand are shown in shades of pink.}
\label{fig:fingerTypeResults-fingers}
\end{figure*}

The effect of handedness on fingerprint quality and performance was already considered in a 2010 work carried out on a database of 40 subjects, evenly distributed between right- and left-handed \cite{Mershon10fpQhandedness}. In that study, both the index and middle fingers of both hands were acquired with three scanners of different technologies (optical, thermal and capacitive). Due to the scarcity of data, results did not show any conclusive trends regarding the impact of handedness on fingerprint performance. The present findings amend the final observations of \cite{Mershon10fpQhandedness} and support the assumption that there is a difference in accuracy between the use of the dominant and non-dominant hand for fingerprint recognition.

While the results presented in this work indicate that handedness has an impact on fingerprint quality, such a statement still needs to be verified with a specific experimental protocol where information regarding handedness is part of the available metadata for the subjects in the dataset (which was not the case for the current study).

\textbf{HYPOTHESIS 1}. When touch-based slap acquisition readers are used, the dominant hand produces higher quality fingerprint impressions than the non-dominant hand.

\textbf{SOLUTIONS}. Assuming the correctness of the hypothesis above, in systems requiring fingers of only one hand, priority could be given to the dominant one. In general it would be useful to store as part of the personal data of individuals, their handedness, so as to give preference, for recognition purposes, to fingerprints from their dominant hand.

In Fig.\ref{fig:fingerTypeResults-fingers}, we present the box plots of the quality distributions for each individual finger of the left and right hands, following the natural order of the fingers of both hands.

\textbf{OBSERVATION 2}. From this figure it can be seen that the quality of fingerprints differs quite significantly depending on the finger-type. Fingers, ordered from lower to higher quality of their fingerprints are: little, ring, middle, index and thumb. This is consistent across both hands. This order follows the natural anatomical order of fingers in the hand.
The most noticeable difference among all fingers, is the low quality produced by the little finger compared to all other four.

\textbf{DISCUSSION}. In their natural state, it is likely that all fingerprints present a similar amount of discriminative information (``character'' definition of quality in ISO/IEC 29794- 1), however, due to the ergonomics/usability of scanners, this information is better captured for some fingers (``fidelity'' definition of quality in ISO/IEC 29794-1).

Touch-based slap acquisition devices require the user to press all four fingers contemporarily against a flat platen, following a straight forward line from the subject. From an anatomical perspective, due to the limitations of the wrist and finger joints, this task is easier to perform with the index finger, and becomes increasingly less comfortable for the rest of the fingers. The result is a good interaction of the index finger with the scanner, that worsens successively for the other fingers. This diverse interaction, in turn, results in different quality of the captured images.

Another factor to be taken into account is that, when capturing all four fingers contemporarily, it is more difficult for the subject to control placing and the amount of pressure exerted by each single one of them separately. It has been pointed out in different works that, when using touch-based scanners, the pressure applied against the platen is one of the key parameters that determines the final quality level obtained for the resulting fingerprint images \cite{Kukula09fpQpressure}.

Following a similar rationale, since thumbs are acquired on their own, subjects are more proficient at placing them correctly on the platen, also having a better control over the pressure applied to each of them, with independence of the rest of fingers, resulting in high quality images. These results confirm, on a statistically significant database, what was initially pointed out by the two works presented in \cite{Gafurov10fingerTypePerformance} and \cite{Michels10fingerCombinationsFpRec}.

\textbf{HYPOTHESIS 2}. The difference in quality among individual fingers is due, not to the distinctiveness of the natural fingerprints, but to the way in which the information contained in them is translated to the digital domain by current touch-based slap acquisition scanners.

\textbf{SOLUTIONS}. While we cannot be certain regarding the cause of the fidelity issue detected in the experiments for the different individual fingers, we can hypothesise that such a difference between character and utility may be put down to a large extent to the ergonomics and usability of current touch-based scanners for the acquisition of slaps impressions. While the performance of current readers is very high, it could still be improved to better acquire the ring and little fingers. This could be accomplished, for instance, by not using a flat platen, but rather a slightly curved one, for example in the shape of a dome (or the top segment of a half sphere). Another possibility would be to change the angle of the flat platen, not to be perpendicular to the body, but to capture the fingers following the natural angle formed at the elbow by the arm and the forearm when it is comfortable rested on a desk or when the hand is placed in front of the chest with the elbow bent.
From a general perspective, further investment should be dedicated to the development of scanners (e.g., researching in potential benefits of touchless technology), the improvement of acquisition protocols (e.g., placement of the scanners, type of fingers to be captured), and the improvement of the usability of this technology both by operators and captured subjects.

If all four fingers are being acquired, from a quality perspective, it is preferable to acquire them individually one by one, rather than as a slap image (all four simultaneously). This way, each finger interacts with the capturing device independently of the other fingers, which would allow the subject to have better control over the acquisition process and, in turn, it would likely improve the quality of fingerprints, especially for the ring and little fingers. Of course, this would entail a significant increase in the acquisition time, which can be a critical factor in some practical applications/scenarios.

For some specific applications it may not possible to acquire all 10 fingers, or it may be decided due to different constraints not to acquire them (e.g., restricted acquisition time). In these cases where an \textit{a-priori} decision must be taken regarding which individual fingers to acquire, priority should be given to, in this order: thumb, index, middle, ring and little fingers.

Recognition algorithms may also make use of this a priory knowledge regarding the expected quality of fingerprints depending on the finger that produced them. For instance, specific score-level fusion strategies could be designed in order to give a higher weight in the final comparison outcome to those fingers that are known to provide better quality \cite{Poh12frameworkQfusion}.

\section{Conclusion}
\label{sec:conc}

Given the paramount importance of quality in biometrics, a very significant amount of effort has been dedicated from all stakeholders in the field (researchers, practitioners, users, developers, vendors) to study the main factors that have an impact on the quality of different biometric characteristics. In particular, as happens in many other areas related to biometrics, fingerprints stand out as the biometric characteristic, where the largest amount of research and information has been generated. In fact, the big investment made in fingerprint quality assessment, has paved the way for other biometric characteristics, such as face, to get the support required in order to reach a similar level of development in terms of understanding of quality.

However, even if undeniable progress has been achieved in fingerprint quality analysis, there are still areas where further research needs to be performed in order to confirm or complement some preliminary observations that have been made on statistically limited sets of data. The present paper is a contribution to bring further insight to this field and to bridge some of these still existing gaps.

In particular, the present work is focused on determining the impact that different demographic factors have on the quality and overall performance of fingerprints in automated recognition systems. The experimental analysis has aimed at assessing the potential bias that may exist in fingerprint recognition technology with respect to gender and age, and, also, to the inter-dependency of fingerprint accuracy with regard to handedness and finger-type.

The results reached in the work, performed on a database of almost 16,000 subjects acquired under real operational conditions, can lead to practical decisions for the improvement on the use and deployment of this technology.
To sum-up the main contributions of the work, the next concrete observations and follow-up experimentation have been extracted from the results presented in the paper:

GENDER:
\begin{itemize}
\item OBSERVATION: there exists a bias in fingerprint recognition between men and women. Fingerprints coming from men, systematically produce higher quality levels and, as a result, also higher accuracy, than those of women.
\item EXPLANATION: such bias is hypothesised to be produced by the difference in ridge density between male and female fingerprints.
\item FOLLOW-UP ACTIONS: higher resolution readers, and specific processing algorithms for fingerprints with a higher ridge density (those of women) may contribute to reduce the observed bias. Further experimentation is required to confirm or reject this hypothesis.
\end{itemize}

AGE:
\begin{itemize}
\item OBSERVATION: there exists a bias in fingerprint recognition for adults and elders over 50 years of age. Fingerprint quality starts decreasing linearly at around 45-50 years of age, and can pose a real challenge in terms of error rates for elders over 65.
\item EXPLANATION: this quality and accuracy difference between young adults and elders, which is consistent both for men and women, is assumed to be produced by the degradation of the skin properties (mainly elasticity and dryness due to loss in collagen levels) which results in an inadequate interaction with the touch-based technology used today in most cases for fingerprint acquisition.
\item FOLLOW-UP ACTIONS: the use of the new generation of touchless fingerprint scanners can help to reduce the observed bias for elders. As in the case of gender, further specific experimentation is required to verify such hypothesis.
\end{itemize}

HANDEDNESS and FINGER-TYPE:
\begin{itemize}
\item OBSERVATION: there exists a bias between the quality provided by fingerprints of the right hand and the left hand, and also among individual fingers, being the most accurate the thumb and index, and the one providing clearly the lowest quality the little finger.

\item EXPLANATION: the explanation for these observed differences is a combination of: 1) handedness: individuals provide better fingerprint quality with their dominant hand; and 2) ergonomics: touch-based slap fingerprint readers are designed to better capture thumbs and indexes, and less adapted for the acquisition of the ring and little finger.

\item FOLLOW-UP ACTIONS: Working on the design, from an ergonomic and usability perspective, of fingerprint slap scanners can help to reduce the quality and accuracy difference between hands and also among individual fingers.
\end{itemize}

To conclude, we can state that the present study has shown that there exist biases among the fingerprint quality of different demographic groups and finger-types. The question that we should ask ourselves is: is this bias due to intrinsic differences in the amount of information contained in our natural fingers? Do fingers from elders contain less information than those of adults? Do fingers from women contain less information than those of men? Is the little finger worse suited for recognition purposes than the index? Or rather, are we not being able to translate or digitally capture with enough accuracy, the information contained in natural fingers of certain segments of the population into the digital domain? That is, are fingerprint readers and fingerprint processing algorithms better designed to perform well with certain demographic groups? Or put in another way, considering the topic from the ISO/IEC 29794-1 standard definition of quality: are these discrepancies caused by inherent ``character'' differences of by ``fidelity'' issues originated at the time of acquisition?

Assuming that the observed quality variability is mainly caused by an inadequate acquisition of fingerprints, and not by the fingers themselves, how can we ensure that such bias is minimised and that we ensure uniform, consistent quality and accuracy across all demographic groups and finger-types?

The present study provides some plausible hypotheses for this observed variability, and finally proposes a course of action that can be followed to minimise the quality difference and, ultimately, to improve recognition accuracy.

To sum up the conclusions of the work, we can state that, in order to improve the overall quality of fingerprint recognition technology across all segments of the population, making its accuracy as consistent and uniform as possible for all demographic groups (e.g., age or gender), we should focus on the design and development of more advanced fingerprint readers, and of specific processing algorithms, capable of extracting all the information contained in fingerprints in a user-friendly and repeatable manner. That is, we should adapt to the specificities of demographic groups in order to minimise potential performance biases among them.


%

%

\ifCLASSOPTIONcompsoc
  \section*{Acknowledgments}
\else
  \section*{Acknowledgment}
\fi

The authors would like to thank the Swedish Agency for Migration (Migrationsverket) for participating in the pilot project that also allowed the fulfillment of the present research work. Authors would especially like to mention Ulf {\AA}kerberg, for his invaluable collaboration and support.

\ifCLASSOPTIONcaptionsoff
  \newpage
\fi



%

%
%


\bibliographystyle{IEEEtran}
\bibliography{TBIOM24}

\end{document}